\pdfoutput=1
\documentclass[11pt]{article}

\usepackage[]{emnlp2021}

\usepackage{times}
\usepackage{latexsym}

\usepackage[T1]{fontenc}
\usepackage[utf8]{inputenc}

\usepackage{microtype}
\usepackage{hyperref}
\usepackage{amsmath}
\usepackage{amsfonts,amssymb}
\usepackage{bbold}
\usepackage{graphicx}
\usepackage{subfigure}

\title{The NiuTrans System for the WMT21 Efficiency Task}

\author{Chenglong Wang\textsuperscript{1}, Chi Hu\textsuperscript{1}, Yongyu Mu\textsuperscript{1},  Zhongxiang Yan\textsuperscript{1}, Siming Wu\textsuperscript{1},\\
\bf{Minyi Hu\textsuperscript{1}, Hang Cao\textsuperscript{1}, Bei Li\textsuperscript{1}, Ye Lin\textsuperscript{1}, Tong Xiao\textsuperscript{1,2}} and Jingbo Zhu\textsuperscript{1,2}\\

    \textsuperscript{1}NLP Lab, School of Computer Science and Engineering, Northeastern University\\
    \textsuperscript{2}NiuTrans Research, Shenyang, China\\
    \ttfamily{clwang1119@gmail.com,huchinlp@gmail.com}\\
    \ttfamily{\{xiaotong,zhujingbo\}@mail.neu.edu.cn}
}

\begin{document}
\maketitle
\begin{abstract}
This paper describes the NiuTrans system for the WMT21 translation efficiency task\footnote{\url{http://statmt.org/wmt21/efficiency-task.html}}. Following last year's work, we explore various techniques to improve efficiency while maintaining translation quality. We investigate the combinations of lightweight Transformer architectures and knowledge distillation strategies. Also, we improve the translation efficiency with graph optimization, low precision, dynamic batching, and parallel pre/post-processing. Our system can translate 247,000 words per second on an NVIDIA A100, being 3$\times$ faster than last year's system. Our system is the fastest and has the lowest memory consumption on the GPU-throughput track.  The code, model, and pipeline will be available at NiuTrans.NMT\footnote{\url{https://github.com/NiuTrans/NiuTrans.NMT}}.
\end{abstract}

\section{Introduction}

Large and deep Transformer models have dominated machine translation (MT) tasks in recent years \cite{vaswani2017attention, edunov2018understanding, wang2019learning, raffel2020exploring}. Despite their high accuracy, these models are inefficient and difficult to deploy \cite{wang2020hat, hu2021ranknas, lin2021bag}. Many efforts have been made to improve the translation efficiency, including efficient architectures \cite{li2020learning, DBLP:journals/corr/abs-2101-00542}, quantization \cite{bhandare2019efficient, DBLP:conf/ijcai/LinLLXLZ20}, and knowledge distillation \cite{li2020shallow, lin2020weight}. 

This work investigates efficient Transformers architectures and optimizations specialized for different hardware platforms. In particular, we study deep encoder and shallow decoder Transformer models and optimize them for both GPUs and CPUs. Starting from an ensemble of three deep Transformer teacher models, we train various student models via sequence-level knowledge distillation (\textsc{Skd}) \cite{hinton2015distilling, li2020learning, kim2016sequence} and data augmentation \cite{shen2020simple}. We find that using a deep encoder (6 layers) and a shallow decoder (1 layer) gives reasonable improvements in speed while maintaining high translation quality. We improve the student model's efficiency by removing unimportant components, including the FFN sub-layers and multi-head mechanism.  We also explore other model-agnostic optimizations, including graph optimization, dynamic batching, parallel pre/post-processing, 8-bit matrix multiplication on CPUs, and 16-bit computation on GPUs. 

Section \ref{sec:model-overview} describes the training procedures of the deep teacher models. Then, Section \ref{sec:various-optimizations} presents various optimizations for reducing the model size, improving model performance and efficiency. Finally, Section \ref{sec:final-submission} details the accuracy and efficiency results of our submissions for the shared efficiency task.

\section{Model Overview}
\label{sec:model-overview}
Following \citet{hu-etal-2020-niutrans}, \citet{li2020learning} and \citet{lin2020weight}, we use the \textsc{Skd} method to train our models. Our experiments also show that the \textsc{Skd} method can obtain better performance than the word-level knowledge distillation (\textsc{Wkd}) method, similar to \citet{kim2016sequence}. Therefore, all of student models are optimized by using the interpolated \textsc{Skd} method \cite{kim2016sequence}, and trained on data generated from the teacher models.

\subsection{Deep Transformer Teacher Models}
\label{sec:model-teacher}
Recently, researchers have explored deeper models to improve the translation quality \cite{wang2019learning,li2020shallow,dehghani2018universal,wang2020training}. Inspired by them, we employ deep Transformers as the teacher models. More specifically, we train three teachers with different configurations, including \textit{Deep-30}, \textit{Deep-12-768}, and \textit{Skipping Sublayer-40}. We also utilize \citet{li2019niutrans}'s ensemble strategy to boost the teachers.

\paragraph{Deep-30 Transformer Model:}
We set the number of encoder layers to 30 in the Transformer model. Other hyper-parameters are identical to the vanilla Transformer.
\paragraph{Deep-12-768 Transformer Model:}
This model modifies the number of encoder layers, hidden sizes and embedding sizes to 12, 3,072 and 768. Such a setting makes the Transformer model deeper and wider. Other hyper-parameters are the same as vanilla Transformer.
\paragraph{Skipping Sublayer-40 Transformer Model:}
This model uses a simple training procedure that samples one streaming configuration in each iteration \cite{li2020learning}. The number of encoder layers is 40, and other setups are the same as \citet{li2020learning}.

We adopt the relative position representation (\textsc{RPR}) \cite{shaw2018selfattention} to further improve the teacher models and set the key's relative length to 8.

\subsection{Lightweight Transformer Student Models}
Although the ensemble teacher model delivers excellent performance, our goal is to learn lightweight models. The natural idea is to compress knowledge from an ensemble into the lightweight model using knowledge distillation \cite{hinton2015distilling}. We employ sequence-level knowledge distillation on the ensemble teacher model described in Section \ref{sec:model-teacher}.

\begin{table}[!t]
\begin{center}
\begin{tabular}{lrr}
\hline
Student Model     & Param. & BLEU \\ \hline
Student-6-6-8 & 96M    & 33.2 \\
Student-6-1-8 & 42M    & 33.0 \\
Student-6-1-1 & 42M    & 32.9 \\ \hline
\end{tabular}
\end{center}
\caption{Reference BLEU scores for the student models on \textit{newstest20}. 6-6-8 means that the model contains 6 encoder layers and 6 decoder layers with 8 attention heads. Other  hyper-parameters  are  the  same  as the vanilla Transformer.}
\label{tab:student-comparsion}
\end{table}

\paragraph{Seqence-level Knowledge Distillation}
The \textsc{Skd} will make a student model mimic the teacher's behaviors at the sequence level. Moreover, the method considers the sequence-level distribution specified by the model over all possible sequences $\mathbf{t}\in T$. Following \citet{kim2016sequence}, the loss function of \textsc{Skd} method for training students is 
\begin{eqnarray}
\mathcal{L}_{\text {SKD}} & \approx & -\sum_{\mathbf{t} \in \mathcal{T}} \mathbf{1}\{\mathbf{t}=\hat{\mathbf{y}}\} \log p(\mathbf{t} \mid \mathbf{s}) \\ 
& = & -\log p(\mathbf{t}=\hat{\mathbf{y}} \mid \mathbf{s})
\end{eqnarray}
where $\mathbf{1}\{\cdot\}$is the indicator function, $\hat{\mathbf{y}}$ is the output of teacher model using beam search, $ \mathbf{s} $ symbolizes the source sentence and $p(\cdot|\cdot)$ denotes the conditional probability. We use the ensemble teacher model to generate multiple translations of the raw English sentences. In particular, we collect the 5-best list for each sentence against the original target to create the synthetic training data. However, we select only 12 million synthetic data to train our student models to reduce training costs. We find that student models will not have better performance when increasing the number of training data.

\begin{table*}[!t]
\begin{center}
\begin{tabular}{l|rrrr|rrr}
\hline
Student Model            & N-Enc & Dim-FFN & Param. & Speedup & newstest18 & newstest19 & newstest20 \\ \hline
Student-12-1-512 & 12    & 512 & 56M      & 2.0x       & 45.3       & 41.7       & 33.2       \\
Student-6-1-512  & 6     & 512 & 38M      & 2.3x       & 44.5       & 41.0         & 32.7       \\
Student-6-1-0    & 6     & 0   & 37M      & 2.4x       & 43.9       & 40.6       & 32.4       \\
Student-3-1-512  & 3     & 512 & 28M      & 2.6x       & 42.8       & 40.0         & 31.5       \\ \hline
\end{tabular}
\end{center}
\caption{N-Enc is the number of encoder layers and Dim-FFN denotes the feed-forward network (FFN) size. The Speedup and BLEU results are measured on a TITAN V GPU. The Speedup is calculated comparing with our ensemble teacher model. The student model has not FFN component in the decoder when the Dim-FFN is 0. Evaluation is performed without inference optimizations and with a beam size of 1.}
\label{tab:student-results}
\end{table*}
%table teacher models result
\begin{table}[!t]
\begin{center}
\begin{tabular}{lrr}
\hline
Teacher Model       & Param. & BLEU \\ \hline
Deep-30       & 138M   & 32.8 \\
Deep-12-768   & 170M   & 33.3 \\
Skipping Sublayer-40 & 171M   & 33.1 \\
Ensemble            & 479M   & 33.4 \\ \hline
\end{tabular}
\end{center}
\caption{Results on \textit{newstest20}-Teacher Models. We train our teacher models with the \textsc{Rpr} and back-translation.}
\label{tab:teacher-results}
\end{table}
\paragraph{Fast Student Models}
\label{sec:fast-student}
As suggested in \citet{hu-etal-2020-niutrans}, the bottleneck of translation efficiency is the decoder part. Hence, we accelerate the decoding by reducing the number of decoder layers and removing multi-head mechanism\footnote{Although the multi-head mechanism does not increase the parameter of the model, it brings non-negligible computational costs.}. Inspired by \citet{hu2021ranknas}, we design the lightweight Transformer student model with one decoder layer. We further remove the multi-head mechanism in the decoder's attention modules. Table \ref{tab:student-comparsion} shows that the Transformer student model with fewer decoder layers and decoder attention heads can achieve similar translation quality to the baseline. Therefore, we train four different student models based on the Transformer architecture with one decoder layer and a single decoder attention head. Those student models are described in detail in Table \ref{tab:student-results}. Besides, experiments show that adding more encoder layers cannot improve the performance when the student model has 12 encoder layers. Therefore, our submissions have 12 encoder layers at most.

\subsection{Data and Training Details}
Our data is constrained by the condition of the WMT 2021 English-German news translation task\footnote{\url{https://www.statmt.org/wmt21/translation-task.html}}, and we use the same data filtering method as \citet{zhang2020niutrans}. We select 20 million pairs to train our teacher models after filtering all official released parallel datasets (without official synthetic datasets). The data is tokenized with Moses \cite{koehn2007moses}, and jointly Byte-Pair Encoded (BPE) \cite{sennrich2015neural} with 32K merge operations using a shared vocabulary. After decoding, we remove the BPE separators and de-tokenize all tokens with Moses \cite{koehn2007moses}.

\paragraph{Teacher Models Training}
We train three teacher models using \textit{newstest19} as the development set with \texttt{Fairseq} \cite{ott2019fairseq}. We share the source-side and target-side embeddings with the decoder output weights. We use the Adam optimizer \cite{kingma2014adam} with $\beta_{1}=0.9$, $\beta_{2}=0.997$ and $\epsilon=10^{-8}$ as well as gradient accumulation due to the high GPU memory footprints. Each model is trained on 8 TITAN V GPUs for up to 11 epochs. The learning rate is decayed based on the inverse square root of the update number after 16,000 warm-up steps, and the maximum learning rate is 0.002. After training, we average the last five checkpoints in the training process for all models. Similar to \citet{zhang2020niutrans}, we train our teacher models with a round of back-translation with 12 million monolingual data selected from the News crawl and News Commentary. We train three De$\rightarrow$En models with the same method and model setup to generate pseudo-data. Table \ref{tab:teacher-results} shows the results of all teacher models and their ensemble, where we report SacreBLEU \cite{post2018call} and the model size. Our final ensemble teacher model can achieve a BLEU score of 33.4 on \textit{newstest20}.

\paragraph{Student Models Training}
The training settings for student models are the same for the teacher models, except its learning rate is 7e$^{-4}$ and warmup-updates are 8,000. In addition, we also use the cutoff method \cite{shen2020simple} to boost our student models\footnote{\url{https://github.com/stevezheng23/fairseq_extension/tree/master/examples/translation/augmentation}} and we train our student model with 21 epochs. Table \ref{tab:student-results} shows the results of all student models. Our student model yields a significant speedup (2$\times$-2.6$\times$) with modest sacrifice in terms of BLEU (0.2-0.9 on \textit{newstest20}).

\subsection{Interpretation of Results}
After training the final student models, we evaluate their BLEU scores on the English-German \textit{newstest20}, \textit{newstest19}, and \textit{newstest18} before any inference optimization. Results show that the student models can achieve very similar performance to the teachers. For instance, the \textit{Student-12-1-512} model delivers a loss of 0.2 BLEU score compared to the ensemble of teacher models. 

\section{Optimizations for Decoding}
\label{sec:various-optimizations}
Our optimizations for decoding are implemented with NiuTensor \footnote{\url{https://github.com/NiuTrans/NiuTensor}}. The optimizations can be divided into three parts, including optimizations for CPUs, GPUs, and device-independent techniques.

\subsection{Optimizations for GPUs}
For the GPU-based decoding, we mainly explore dynamic batching and FP16 inference.
\paragraph{Dynamic Batching}
Unlike the CPU version, the easiest way to reduce the translation time on GPUs is to increase the batch size within a specific range. We implement a dynamic batching scheme that maximizes the number of sentences in the batch while limiting the number of tokens. This strategy significantly accelerates the inference compared to a fixed batch size when the sequence length is short. 
\paragraph{FP16 Inference} 
Since the Tesla A100 GPU supports calculations under FP16, our systems execute almost all operations in 16-bit floating-point. To escape overflow, we convert the data type before and after the softmax operation in the attention modules. We also reorder some operations for numerical stability. For instance, we apply the scaling operation (dived by $\sqrt{d_k}$) to the query instead of the attention weights. To accelerate our systems further, we replace the vanilla layer normalization with the L1-norm \cite{DBLP:conf/ijcai/LinLLXLZ20}. Also, we find that removing the multi-head mechanism (by setting the head to 1) in the student models significantly improves the throughput without performance loss.

\begin{figure}[t!]
    \centering
    \includegraphics[scale=0.4]{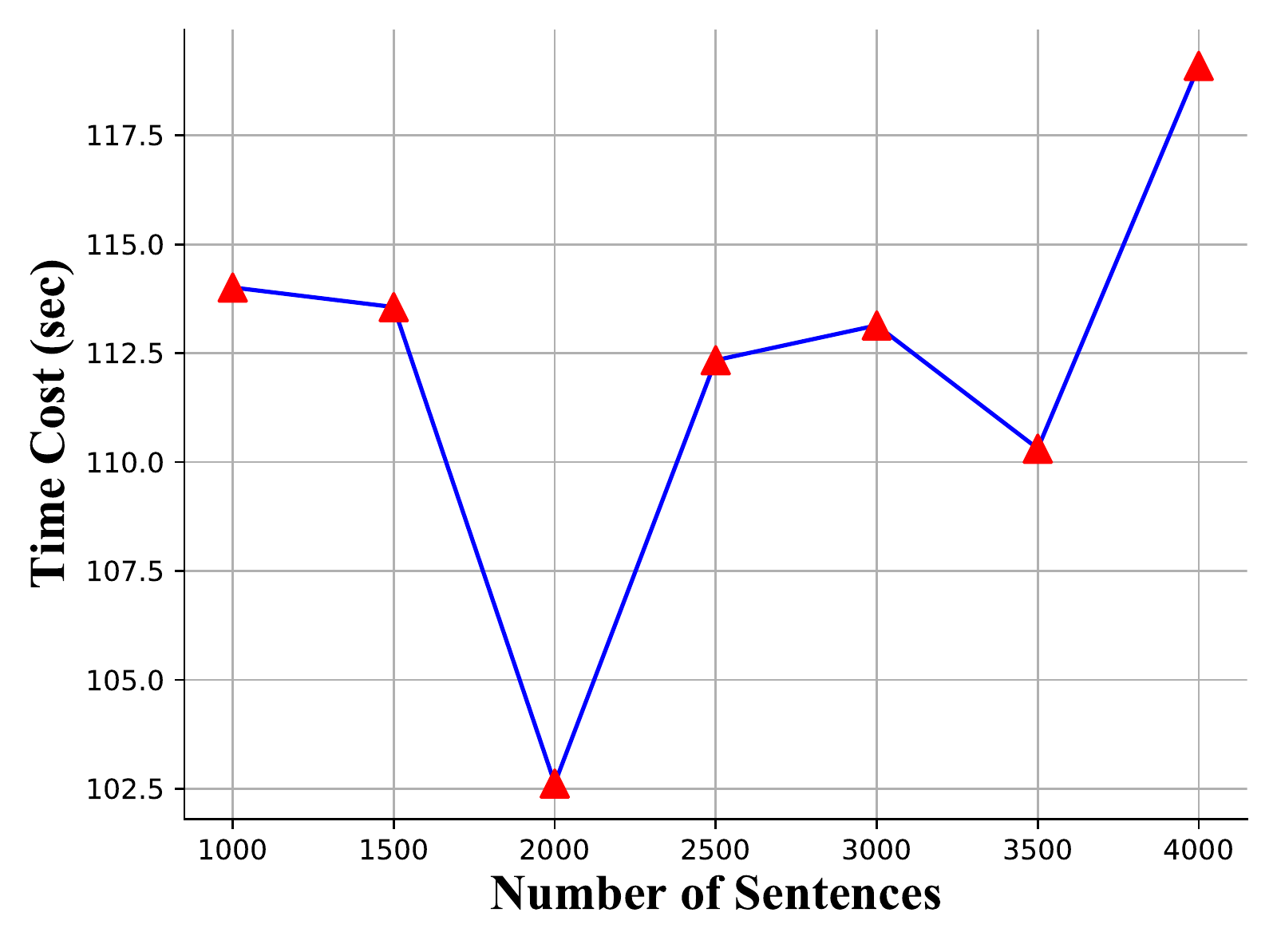}
    \caption{Results on Student-6-1-512 model. The time cost is measured on an Intel Xeon Gold 6240 CPU with 100,000 lines of raw English sentences with an averaged length of 18 words.}
    \label{fig:sent-num}
\end{figure}

\begin{figure*}[t!]
    \centering
    \subfigure[Performance of our GPU system]{
    % \begin{minipage}[t]{1\linewidth}
    \centering
    \includegraphics[scale=0.43]{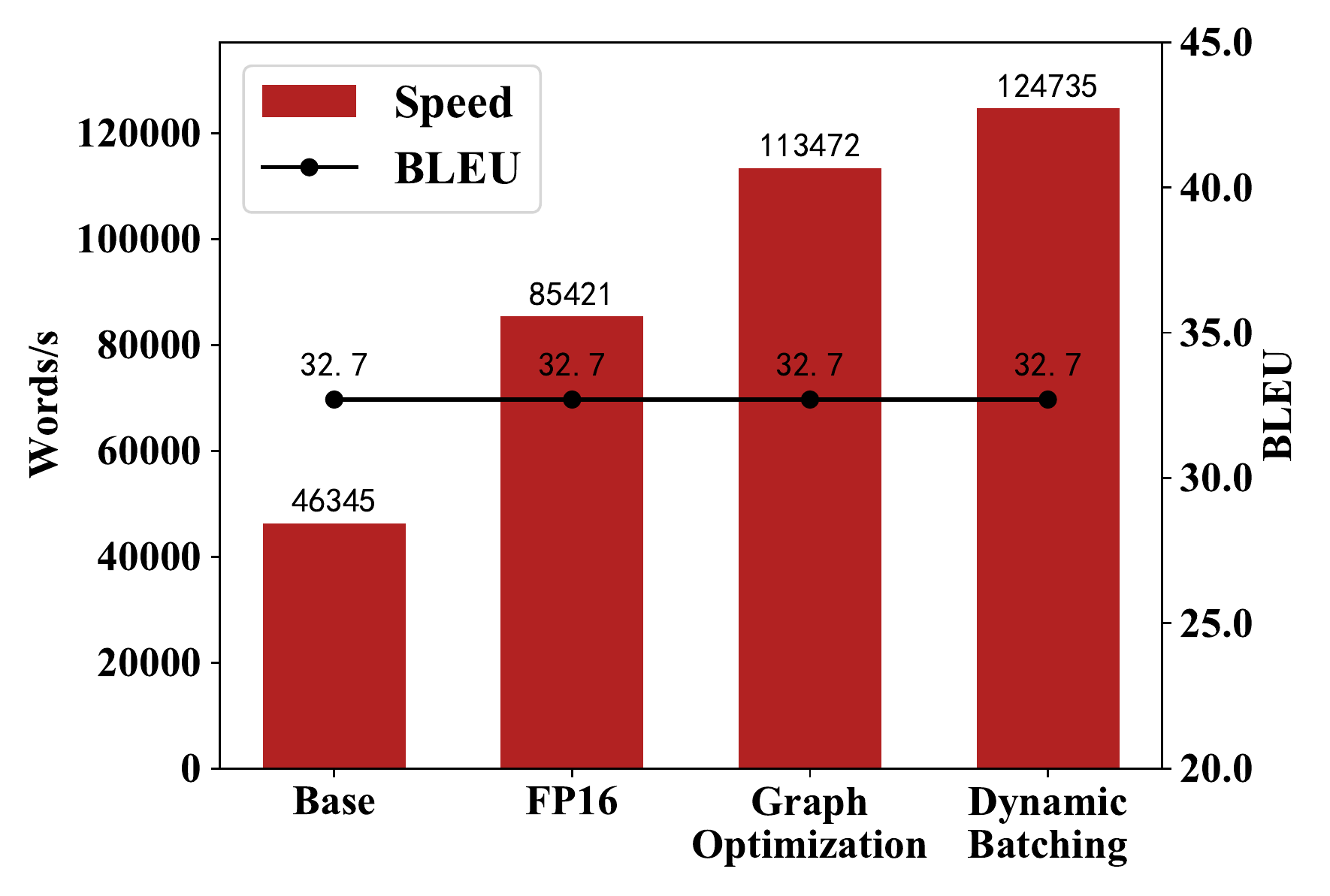}
    %\caption{fig1}
    % \end{minipage}%
    \label{fig:final-bleu}
    }%
    \subfigure[Performance of our CPU system]{
    % \begin{minipage}[t]{1\linewidth}
    \centering
    \includegraphics[scale=0.43]{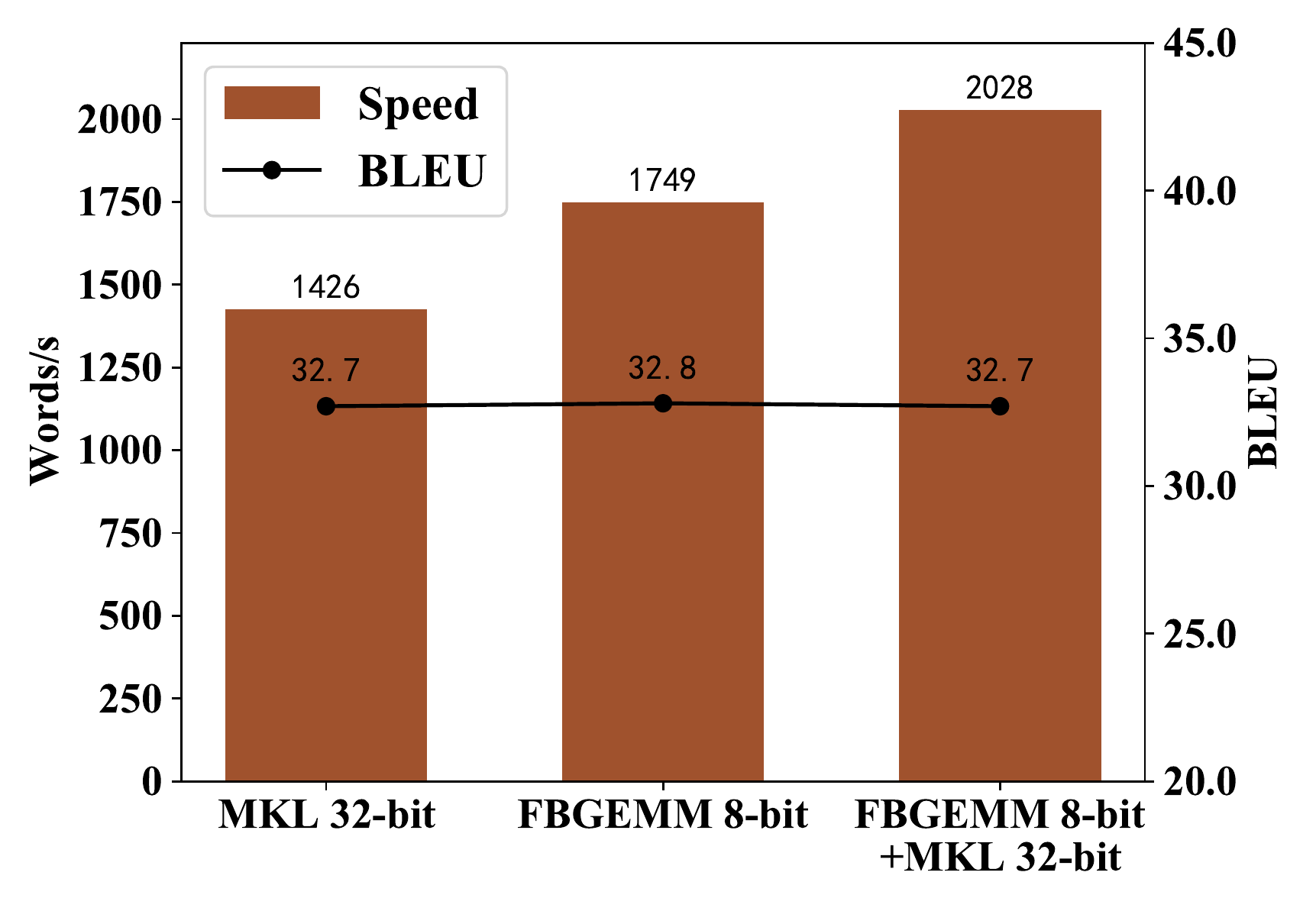}
    %\caption{fig2}
    % \end{minipage}%
    \label{fig:final-speed}
    }%
    \caption{BLEU on \textit{newstest20} versus words per second (Words/s) with different optimizations on a TITAN V GPU and Intel Xeon Gold 5118 CPUs. Result of decoding speed is measured with 0.1 M sentences (average length is 18). When the GPU system is running, it will use all free CPUs on the device. }
    \label{fig:final-results}
\end{figure*}

\subsection{Optimizations for CPUs}
We employ the \textit{Student-6-1-512} and \textit{Student-3-1-512} models as our CPU submissions. Two methods are discussed to speed up the decoding for our CPU systems.

\paragraph{The Use of MKL}
We use the Intel Math Kernel Library \cite{wang2014intel} to optimize our NiuTensor framework, which helps our systems to make the full use of the Intel architecture and to extract the maximum performance.

\paragraph{8-bit Matrix Multiplication with Packing}
We implement 8-bit matrix multiplication using the open-source library FBGEMM \cite{khudia2021fbgemm}. Following \citet{kim2019research}, we quantize each column of the weight matrix separately with different scales and offsets. Scale and offsets for weight matrix are calculated by:

\begin{equation}
b_{scale}[j]=\frac{14\sigma _{j}}{255} 
\end{equation}
\begin{equation}
b_{zeropoint}[j]=\frac{127-(\bar{x}_{j}+7\sigma_{j})}{b_{scale}[j]}
\end{equation}
where $\sigma _{j}$ and $\bar{x}_{j}$ refer to average and standard deviation for the $j$-th column. The quantization parameters for the input matrix is calculated by:
\begin{equation}
a_{scale}=\frac{x_{max}-x_{min}}{255} 
\end{equation}
\begin{equation}
a_{zeropoint}=\frac{255-x_{max}}{a_{scale}}
\end{equation}
where $x_{max}$ and $x_{min}$ are the maximum and minimum values of the matrix respectively. With FBGEMM API, we also execute the packing operation to change the layout of the matrices into a form that uses the CPU more efficiently. We pre-quantize and pre-pack all the weight matrices to avoid repeated operation during inference.

\subsection{Other Optimizations}
Furthermore, we explore other device-independent methods to optimize our systems. Those methods help our systems to achieve obvious speed-up without translation precision loss.

\paragraph{Graph Optimization}
Furthermore, we explore other device-independent methods to optimize our systems. Those methods help our systems to achieve significant speed-up without translation precision loss.
\begin{itemize}
    \item Computation optimization. We prune all redundant operations and reorder some operations in the computational graph. For instance, we remove the \textit{log-softmax} operation in the output layer when using greedy search. We also extract the \textit{transpose} operations from matrix multiplications to the beginning of decoding.
    \item Memory optimization. We reuse all possible nodes to minimize memory consumption. We also reduce the memory allocation or movement with an efficient memory pool. Moreover, we sort the source sentences in descending order of length and detect the peak memory footprint before decoding.
\end{itemize}

\paragraph{Parallel Execution}
We use the GNU Parallel \cite{Tange2011a} for our systems to perform tasks in parallel. More specifically, we split the standard input into several lines and deliver them via the pipeline. The method is used to accelerate pre-processing, post-processing, and decoding on CPUs. We also find that the system decoding speed/memory is strongly correlated with the number of lines per task. To find the best number of lines for each run, we measure the time cost in different setups against the number of lines. Figure \ref{fig:sent-num} shows that 2,000 is a relatively good choice, and the \textit{Student-6-1-512} model can translate 100,000 sentences in 102.6s on CPUs under this setup.

\paragraph{Better Decoding Configurations}
As aforementioned, our GPU versions use a large batch size, but the batch size on the CPU is much smaller. We use (\textit{sbatch}) and (\textit{wbatch}) to restrict the number of sentences and number of words in a mini-batch not to be greater than \textit{sbatch} and \textit{wbatch}, respectively. In our GPU systems, we set the \textit{sbatch}/\textit{wbatch} to 3,072/64,000. For our CPU systems, the number of processes is managed by the Parallel tool, which is more efficient and accurate. Moreover, We use one MKL thread for each process and set the \textit{sbatch}/\textit{wbatch} to 128/2,048. 

\paragraph{Greedy Search}
In the practice of knowledge distillation, we find that our systems are insensitive to the beam size. It means that the translation quality is good enough even using greedy search in all submissions.

\paragraph{Fast Data Preparation}
We use the fastBPE\footnote{\url{https://github.com/glample/fastBPE}}, a faster C++ version of subword-nmt\footnote{\url{https://github.com/rsennrich/subword-nmt}}, to speed the BPE process. Moreover, we also use the fast-mosestokenizer\footnote{\url{https://github.com/mingruimingrui/fast-mosestokenizer}} for tokenization.

\subsection{Results after Optimizations}
Figure \ref{fig:final-results} plots the \textit{Student-6-1-512} model’s performance with different decoding optimizations. All results show that our optimizations can significantly speed up our system without losing BLEU. Interestingly, we observe additional improvements of 0.4/0.1 BLEU points on the GPU/CPU through decoding optimizations in all our experiments. We also measure other models after decoding optimizations and find their performance is similar to the \textit{Student-6-1-512} model.

\section{Submissions and Results}
\label{sec:final-submission}
\subsection{Submissions}
Our GPU submissions are compiled with CUDA 11.2. We set the number of decoder layers, and the number of decoder attention heads to one as described in Section \ref{sec:fast-student} for all our GPU systems. \textit{Student-12-1-512} model gives a speedup of more than 6$\times$ on the GPU with a slight decrease of 0.2 BLEU on the \textit{newstest20} compared to the deep ensemble model. The system is named as \textbf{\textit{Base-GPU-System}} in following part. We continue to reduce the number of encoder layers for more accelerations. The GPU system with \textit{Student-6-1-512} model improves the translation speed by 25\% with 6 less encoder layers compared to the \textit{Base-GPU-System}. Our fastest GPU system consists of three encoder layers and one decoder layer, which achieves 31.5 BLEU on the \textit{newstest20} with GPU and 1.6$\times$ speedup compared to the \textit{Base-GPU-System}. We also employ the \textit{Student-6-1-0} model to create a GPU system that can achieve the 1.3$\times$ speedup compared to \textit{Base-GPU-System}. Our systems are compiled in the 11.2.1-devel-centos7 docker image, an NVIDIA open-source image\footnote{\url{https://hub.docker.com/r/nvidia/cuda}}. We copy the executable, dependence tools, and model files to the 11.2.1-base-centos7 docker images (final submission). In this way, we ensure all of our system docker images can be executed by the organizers successfully and reduce the docker images size.

For the CPU track submissions, we use the test machine, which has 18 virtual cores. Our CPU version is compiled with MKL static library, and the executable file is 23MiB. Also, we use the 8-bit matrix multiplication with packing to speed the matrix multiplication in the network. We use the \textit{Student-3-1-512} and \textit{Student-6-1-512} models in our CPU systems, and they respectively achieve 31.5 and 32.8 BLEU on \textit{newstest20}. For our CPU docker images, we use the base-centos7 docker image\footnote{\url{https://hub.docker.com/_/centos}} to deploy our CPU MT systems.

Furthermore, all submissions are tested with different cases, including dirty data, empty input, and very long sentences. The test results show that our systems can run successfully with exceptional inputs.

\subsection{Results}
Our systems for the GPU-throughput track are the fastest overall submissions. Specifically, the \textit{Student-3-1-512} system can translate about 250 thousand words per second and achieve 25.5 BLEU on \textit{newstest21}. We attribute this to the comparison of the performance of our teacher model on WMT21. In the CPU track, our system also has competitive performance. Our fastest CPU system created by \textit{Student-3-1-512} model can translate about 48 thousand words per real second via 36 CPU cores and can achieve 25.5 BLEU. We find that reducing the number of encoder layers for student model achieves lower BLEU scores at a similar speed for our CPU systems. Moreover, we compare the cost-efficiency of GPU and CPU decoding in terms of millions of words translated per dollar according to the official evaluation results. We find that translating on GPUs is much more cost-effective than on CPUs. Notably, our GPU system with the \textit{Student-3-1-512} model can translate 300 million words per dollar with acceptable quality. Also, all of our GPU systems have the lowest RAM consumption (about 4 GB) over all submissions according to official test.

\section{Conclusion}
We have described our systems for the WMT21 shared efficiency task. We have explored various efficient Transformer architectures and optimizations specialized for both CPUs and GPUs. We have shown that a lightweight decoder and proper optimizations for different hardware can significantly accelerate the translation process with slight or no loss of translation quality. Our fastest GPU system with three encoder layers and one decoder layer is 11$\times$ faster than the deep ensemble model and lose 1.9 BLEU points. 

\section*{Acknowledgements}
This work was supported in part by the National Science Foundation of China (Nos. 61876035 and 61732005), the National Key R\&D Program of China (No.2019QY1801), and the Ministry of Science and Technology of the PRC (Nos. 2019YFF0303002 and 2020AAA0107900). The authors would like to thank the anonymous reviewers for their comments and suggestions.

\bibliography{anthology,custom}
\bibliographystyle{acl_natbib}

\appendix

% \section{Example Appendix}
% \label{sec:appendix}

\end{document}